# Derivations of Normalized Mutual Information in Binary Classifications

Yong Wang, Bao-Gang Hu, *Senior Member, IEEE*

*Abstract*—This correspondence studies the basic problem of classifications - how to evaluate different classifiers. Although the conventional performance indexes, such as accuracy, are commonly used in classifier selection or evaluation, information-based criteria, such as mutual information, are becoming popular in feature/model selections. In this work, we propose to assess classifiers in terms of normalized mutual information (NI), which is novel and well defined in a compact range for classifier evaluation. We derive close-form relations of normalized mutual information with respect to accuracy, precision, and recall in binary classifications. By exploring the relations among them, we reveal that NI is actually a set of nonlinear functions, with a concordant power-exponent form, to each performance index. The relations can also be expressed with respect to precision and recall, or to false alarm and hitting rate (recall).

## I. Introduction

Model evaluation or selection [1], [2] is the key to making real progress in machine learning. There are so many widely used or recently developed machine learning methods which form different learning models (The reader is referred to [3]-[6] for a comprehensive review of methods of machine learning.), that we need systematic ways to evaluate how different methods work and to compare one with another to determine which ones to use on a particular problem. However, how to choose the model evaluation criteria is still a controversial problem, as there are many different model selection criteria and the relations among them are nebulous.

For pattern recognition, it is natural to measure a classifier's performance in terms of "accuracy", but in view of some specific cases there are still some other classifier selection indexes, such as precision, recall, precision-recall curves [7]-[9], receiver operating characteristic (ROC) curves [10], [11], area under the ROC curve (AUC) [12], [13] etc., which are named as performance-based criteria. Furthermore, there exists another category of criteria named as information-based criteria, such as AIC [14], BIC [15], NIC [16], minimum relative-entropy (also called minimum cross-entropy) [17], maximum entropy [18] etc., and in recent years, some new information-based criteria,, such as Bayesian Entropy Criterion (BEC) [19], "empirical entropy plus penalty term" criterion [20], mutual information [21], [22] etc., are becoming increasing popular and provide improved classification accuracy. Fano [23] and Hellman [24] made fundamental progress on studying their relations and discovered that the upper bond and the lower bond of Bayesian accuracy can be expressed by the combination of entropy and mutual information.

In this work, we propose to assess classifiers in terms of the Normalized Mutual Information criterion (NI), which is a novel and well defined in a compact range for classifier evaluation based on information theory. Though there are different definitions of NI [25]-[28], what we proposed is an asymmetric one, which is used for classifier evaluation under the unifying learning framework of paper [21], [22] and possesses unique merits (those will be elaborated in the next section). Furthermore, we derive close-form relations of NI with respect to accuracy, precision, and recall in binary classifications. By exploring the relations among them, we reveal that NI is actually a set of nonlinear functions, with a concordant power-exponent form, to each performance index. The relations can also be expressed with respect to precision and recall, or to false alarm and hitting rate (recall).

The structure of this paper is as follows: our proposed NI criterion is described in Section 2. Section 3 discusses the

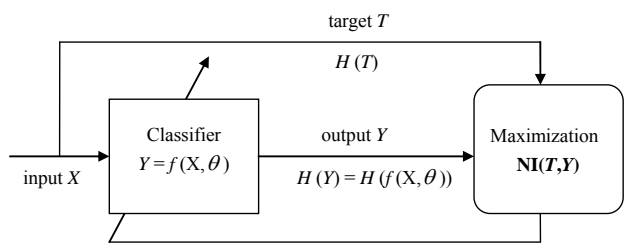

Fig. 1.  Classifier learning model with the maximization normalized mutual information criterion. (Adopted and modified from [21], [22])

Manuscript received November 21, 2007. This work was supported by Natural Science Foundation of China (#60275025, #60121302).

Yong Wang is with the National Laboratory of Pattern Recognition (NLPR), Institute of Automation, Chinese Academy of Sciences (CASIA),  P. O. Box 2728, Beijing 100080, China (phone: +86 10 82614506; fax: +86 10 62647458; e-mail: wangyong@ nlpr.ia.ac.cn).

Bao-Gang Hu is with NLPR/LIAMA, Institute of Automation, Beijing Graduate School, Chinese Academy of Sciences,  P. O. Box 2728, Beijing 100080, China (e-mail: hubg@ nlpr.ia.ac.cn).
.

nonlinear relations between NI and other criteria. In section 4, we illustrate an example of using NI for model evaluation. Finally, a short discussion section ends the paper.

## II. NORMALIZED MUTUAL INFORMATION CRITERION

Information-based criteria are usually used for feature selection or extraction [5], [29], but researchers [30] have explored the feasibility of applying them to model evaluation directly. A unifying learning scheme based on mutual information criterion has been proposed [21], [22] and proceeding with it, we propose to assess classifiers in terms of NI, which is more objective and less time cost. Fig. 1 shows the block diagram of classifier learning model based on maximization NI.

Suppose $Y = f(X, \theta)$ to be the classifier learning model. $X = \{x_i\}_{i=1}^{w}$, $Y = \{y_i\}_{i=1}^{w}$, and $T = \{t_i\}_{i=1}^{w}$ are the input, output, and target of it respectively ( $x_i \in R^n$, $y_i \in R, t_i \in R$ ). $H(T)$ is the entropy of $T$, $H(Y)$ is the entropy of $Y$, and NI $(T,Y)$ is the normalized mutual information of $T$ and $Y$. For supervised classification, $T$ is usually known and it is the real label of $X$, and $Y$ is the estimated label of $X$. So the visual representation of Fig. 1 is to design nonlinear function $Y = f(X, \theta)$ by adjusting the parameter $\theta$ to make the target $T$ and the output $Y$ own the maximization normalized mutual information. In other words, the target function of the classifier learning model is to maximum the correlation of the target $T$ and the output $Y$.

The definitions of NI are not unified, which include symmetric [23], [26] and asymmetric definitions [24], [25]. In the classifier learning model we prefer to choose the asymmetric definition of NI as below:

*Definition 1: Calculation of Normalized mutual information of T and Y.*

$$\text{NI}(T,Y) = \frac{H(T) - H(T|Y)}{H(T)} \quad (1)$$

*Remark 1:* For a given supervised classification problem, $T$ is usually known. So $H(T)$ is a constant and when evaluating different models what we need to do is just to calculate $H(T|Y)$. Thus it can reduce the extra cost for NI computation. Furthermore, $0 \leq H(T|Y) \leq H(T)$, then the value of NI $(T, Y)$ is bonded between 0 and 1. Thus, when comparing different models for different problems it can provide an objective criterion. Especially when doing statistical evaluations, these merits are prominent. The unique merits of Definition 1 are summarized below:
1) NI takes the entropy of target sequence as an isolated item in NI, which is a constant for a given problem. Thus it can reduce the extra cost for NI computation.
2) NI takes value ranges from 0 to 1, where 0 indicates the classifier is utterly not informative; and 1 indicates the classifier can discriminate the classes. Thus it can provide an objective criterion for model evaluation.

Strictly speaking, NI $(T, Y)$, $H(T)$, and $H(T|Y)$ should be calculated based on the distributions of the random variables $T$ and $Y$, but in practice they are usually unavailable. So we propose to calculate them with frequencies as below:

*Definition 2: Calculation of Empirical Entropy $H_e(T)$.*

Given a data set $D = \{(x_i, t_i) | x_i \in R^n, t_i \in R\}_{i=1}^{w}$, $w$ is the total number of samples. All the samples belong to $K$ different classes and their labels make up the target set $T = \{t_i\}_{i=1}^{w}$. $T_i = \{t_{ij}\}_{j=1}^{w_i}$ ($i$=1, 2… $K$) is the subset of $T$ and represents the set of real labels of samples belonging to class $i$, which consists of $w_i$ samples. Then the empirical entropy $H_e(T)$ is defined as below:

$$\begin{aligned} H_e(T) &= H_e(T_1, T_2, \cdots, T_K) \\ &= H\left(\frac{w_1}{w}, \frac{w_2}{w}, \cdots, \frac{w_K}{w}\right) \\ &= -\sum_{i=1}^{K}\left(\frac{w_i}{w}\log_2\left(\frac{w_i}{w}\right)\right) \end{aligned} \quad (2)$$

*Remark 2:* Though $H(T)$ measures the uncertainty or impurity of target set $T$, it also reflects the character of data set $D$. When all the samples of $D$ belong to the same class $i$, i.e. $w_i = w$ and the sample numbers of all the other classes are zeroes, the data set $D$ are completely certain. Then $H(T) = H(T_i) = 0$ (Define $0\log_2 0 = 0$, because $\lim_{p \to 0} p\log_2 p = 0$ ). When the sample numbers of all the classes are equal, the uncertainty of $T$ reaches the maximum $\log_2 K$. So for the data set $D$ composed of $K$ classes, the entropy of its target set $T$ is confined as below:

$$0 \leq H(T) \leq \log_2 K \quad (3)$$

*Definition 3: Calculation of Empirical Conditional Entropy $H_e(T|Y)$.*

$T = \{t_i\}_{i=1}^{w}$ is the target set of classifier learning model and $Y = \{y_i\}_{i=1}^{w}$ is the output set of classifier learning model. $y_i$ is the estimated label of $x_i$ and it is not necessarily equal to $t_i$. According to their values, the entire estimated labels equal to $j$ form a subset of $Y$. Then $Y$ is divided into $K$ subsets denoted as $Y_j = \{Y_{ij}\}_{i=1}^{K}$ ($j$=1, 2… $K$). $Y_{ij}$ is the subset of $Y_j$ and it is composed of the outputs whose estimated labels are $j$ while the target labels are $i$. $Y_{ij}$ contains $w_{ij}$ samples, and then the empirical conditional entropy $H_e(T|Y)$ is defined as below:

$$\begin{aligned} H_e(T|Y) \\ = H_e(T|Y_1, Y_2, \cdots, Y_K) \end{aligned}$$

$$= H_e\left(T \mid \{Y_{i1}\}_{i=1}^K, \{Y_{i2}\}_{i=1}^K, \cdots, \{Y_{iK}\}_{i=1}^K\right)$$

$$= \sum_{j=1}^K \left(\frac{\sum_{i=1}^K w_{ij}}{w} H_e\left(T \mid \{Y_{ij}\}_{i=1}^K\right)\right)$$

$$= -\sum_{j=1}^K \left(\frac{\sum_{i=1}^K w_{ij}}{w} \sum_{i=1}^K \left(\frac{w_{ij}}{\sum_{i=1}^K w_{ij}} \log_2\left(\frac{w_{ij}}{\sum_{i=1}^K w_{ij}}\right)\right)\right)$$

$$= -\sum_{i=1}^K \sum_{j=1}^K \left(\frac{w_{ij}}{w} \log_2\left(\frac{w_{ij}}{\sum_{i=1}^K w_{ij}}\right)\right) \quad (4)$$

*Remark 3:* $H(T|Y)$ measures the degree that the output set *Y* decreases the uncertainty or impurity of the target set *T*. So if the output of a classifier is closer to the target, the value of $H(T|Y)$ will be smaller. Thus $H(T|Y)$ can also be used as a criterion for model evaluation. But contrary to maximize NI (*T*, *Y*), the target function will be converted to minimize $H(T|Y)$.

### III. RELATIONS

The proposed normalized mutual information criterion is quite different from the traditional criteria for model selection. It evaluates model performance from the aspect of information theory directly, but how it is related with traditional criteria is still not well defined. We study their relations for the binary classification problems and derive close-form relations of NI with respect to accuracy, precision, and recall.

In the two-class case with classes *Positive* and *Negative*, a single prediction has four different possible outcomes shown in Table I. The *True Positive* (*TP*) and the *False Negative* (*FN*) compose the *Positive* class; The *True Negative* (*TN*) and the *False Positive* (*FP*) compose the *Negative* class. *T* is the label set of actual classes and *Y* is the label set of predicted classes.

TABLE I
DIFFERENT OUTCOMES OF A TWO-CLASS PREDICTION

|  | Predicted classes (*Y*) | | |
|---|---|---|---|
|  | Positive | Negative | |
| True Positive (*TP*) | False Negative (*FN*) | Positive ($w_1$) | Actual classes (*T*) |
| False Positive (*FP*) | True Negative (*TN*) | Negative ($w_2$) | |

$w_1 = TP + FN$, $w_2 = FP + TN$ and set $w_1 \geq w_2$.

The accuracy, precision, and recall are defined by *TP*, *FN*, *TN* and *FP* as below:

Accuracy (Acc.): $A = \dfrac{TP + TN}{TP + TN + FP + FN}$  (5)

Precision (Pre.): $P = \dfrac{TP}{TP + FP}$  (6)

Recall or Hitting Rate (Rec.): $R = \dfrac{TP}{TP + FN}$  (7)

False Alarm (F.A.): $F = \dfrac{FP}{FP + TN}$  (8)

Following the Definition 2 we can derive the Empirical Entropy *H (T)* as bellow:

$$H(T) = -\frac{w_1}{w_1 + w_2}\log_2\left(\frac{w_1}{w_1 + w_2}\right) - \frac{w_2}{w_1 + w_2}\log_2\left(\frac{w_2}{w_1 + w_2}\right) \quad (9)$$

For a given problem *H (T)* is a constant and following the Definition 1 and 3 we can derive the normalized mutual information of *T* and *Y* as below:

$$NI(T,Y) = \frac{H(T) - H(T|Y)}{H(T)}$$

$$= \begin{bmatrix}
-\dfrac{TP+FN}{TP+TN+FP+FN}\log_2\left(\dfrac{TP+FN}{TP+TN+FP+FN}\right) \\
-\dfrac{FP+TN}{TP+TN+FP+FN}\log_2\left(\dfrac{FP+TN}{TP+TN+FP+FN}\right) \\
+\dfrac{TP}{TP+TN+FP+FN}\log_2\left(\dfrac{TP}{TP+FP}\right) \\
+\dfrac{FP}{TP+TN+FP+FN}\log_2\left(\dfrac{FP}{TP+FP}\right) \\
+\dfrac{TN}{TP+TN+FP+FN}\log_2\left(\dfrac{TN}{TN+FN}\right) \\
+\dfrac{FN}{TP+TN+FP+FN}\log_2\left(\dfrac{FN}{TN+FN}\right)
\end{bmatrix} / H(T)$$

(10)

By introducing (5) - (9) to (10) we can get the nonlinear relations among NI, accuracy, precision, and recall, and it can be classified into nine cases, of which the first eight are special cases and the last (Case 9) is the normal case. They are listed below:

*Case 1:*
$TP + FP = 0 \Leftrightarrow TP = FP = 0 \Leftrightarrow$ P is invalid, R = 0, 0 < A < 1
$\qquad\qquad NI = 0$

*Case 2:* $TN + FN = 0 \Leftrightarrow TN = FN = 0 \Leftrightarrow R = 1, P = A$
$\qquad\qquad NI = 0$

*Case 3:* $TP + TN = 0 \Leftrightarrow TP = TN = 0 \Leftrightarrow A = P = R = 0$
$\qquad\qquad NI = 1$

*Case 4:*
$FP + FN = 0 \Leftrightarrow FP = FN = 0 \Leftrightarrow A = P = R = 1 \Leftrightarrow P + R = 2PR$
$\qquad\qquad NI = 1$

*Case 5:* $TP = 0, TN \neq 0, FP \neq 0 \Leftrightarrow P = R = 0, 0 < A < 1$

$$NI = \frac{1}{(w_1 + w_2) \cdot H(T)} \begin{bmatrix} (w_1 + w_2)\log_2(A^A) \\ +(A+1)\log_2(w_1 + w_2)^{(w_1+w_2)} \\ -\log_2(w_2^{w_2}) \\ -\log_2(Aw_1 + Aw_2 + w_1)^{(Aw_1+Aw_2+w_1)} \end{bmatrix}$$

*Case 6:*

$TP \neq 0, TN = 0, FN \neq 0 \Leftrightarrow 0 < A, P, R < 1, \frac{1}{P} + \frac{1}{R} = 1 + \frac{1}{A}$

$\Leftrightarrow 0 < A, P, R < 1, A(w_1 + w_2) = \frac{Pw_2}{1-P}$

$\Leftrightarrow 0 < A, P, R < 1, A(w_1 + w_2) = Rw_1$

$NI = \frac{1}{(w_1+w_2) \cdot H(T)} \begin{bmatrix} (w_1+w_2)\log_2(A^A) \\ +(A+1)\log_2(w_1+w_2)^{(w_1+w_2)} \\ -\log_2(w_1^{w_1}) \\ -\log_2(Aw_1+Aw_2+w_2)^{(Aw_1+Aw_2+w_2)} \end{bmatrix}$

$= \frac{1}{(w_1+w_2) \cdot H(T)} \begin{bmatrix} \frac{w_2}{1-P}\log_2(P^P) \\ +w_2\log_2(1-P) \\ +\log_2(w_1+w_2)^{(w_1+w_2)} \\ -\log_2(w_1^{w_1}) \\ -\log_2(w_2^{w_2}) \end{bmatrix}$

$= \frac{1}{(w_1+w_2) \cdot H(T)} \begin{bmatrix} w_1\log_2(R^R) \\ +\log_2(w_1+w_2)^{(w_1+w_2)} \\ -(1-R)\log_2(w_1^{w_1}) \\ -\log_2(Rw_1+w_2)^{(Rw_1+w_2)} \end{bmatrix}$

*Case 7:*
$FP = 0, FN \neq 0, TP \neq 0$

$\Leftrightarrow P = 1, 0 < A, R < 1, A(w_1+w_2) = Rw_1 + w_2$

$NI = \frac{1}{(w_1+w_2) \cdot H(T)} \begin{bmatrix} (w_1+w_2)\log_2(1-A)^{(1-A)} \\ +(2-A)\log_2(w_1+w_2)^{(w_1+w_2)} \\ -\log_2(w_1^{w_1}) \\ -\log_2(w_1+2w_2-Aw_1-Aw_2)^{(w_1+2w_2-Aw_1-Aw_2)} \end{bmatrix}$

$= \frac{1}{(w_1+w_2) \cdot H(T)} \begin{bmatrix} w_1\log_2(1-R)^{(1-R)} \\ +\log_2(w_1+w_2)^{(w_1+w_2)} \\ -R\log_2(w_1^{w_1}) \\ -\log_2(w_1+w_2-Rw_1)^{(w_1+w_2-Rw_1)} \end{bmatrix}$

*Case 8:*
$FP \neq 0, FN = 0, TN \neq 0$

$\Leftrightarrow R = 1, 0 < A, P < 1, A(w_1+w_2) = 2w_1 + w_2 - \frac{w_1}{P}$

$NI = \frac{1}{(w_1+w_2) \cdot H(T)} \begin{bmatrix} (w_1+w_2)\log_2(1-A)^{(1-A)} \\ +(2-A)\log_2(w_1+w_2)^{(w_1+w_2)} \\ -\log_2(w_2^{w_2}) \\ -\log_2(w_2+2w_1-Aw_1-Aw_2)^{(w_2+2w_1-Aw_1-Aw_2)} \end{bmatrix}$

$= \frac{1}{(w_1+w_2) \cdot H(T)} \begin{bmatrix} \frac{w_1}{P}\log_2(1-P)^{(1-P)} \\ +w_1\log_2 P \\ +\log_2(w_1+w_2)^{(w_1+w_2)} \\ -\log_2(w_1^{w_1}) \\ -\log_2(w_2^{w_2}) \end{bmatrix}$

*Case 9:* $TP \neq 0, TN \neq 0, FP \neq 0, FN \neq 0$

$NI = \frac{1}{H(T)} \left\{ \log_2(P+R-2PR) + \frac{1}{P+R-2PR} \begin{bmatrix} P(1-A)\log_2(1-R)^{(1-R)} \\ +R(1-A)\log_2(1-P)^{(1-P)} \\ -PR\log_2(1-A)^{(1-A)} \\ +\log_2(AP+AR-PR-APR)^{(AP+AR-PR-APR)} \\ -\log_2(AP+R-2PR)^{(AP+R-2PR)} \\ -\log_2(AR+P-2PR)^{(AR+P-2PR)} \end{bmatrix} \right\}$

It is obvious that NI is a nonlinear function of accuracy, precision, and recall and it integrates them in a nature way. Furthermore the nonlinear function shows a specific character - it is a concordant power-exponent function. Though the function is a four-dimension function that one can not see its map directly, we can see its projected pictures on accuracy, precision, and recall. Fig.2 - Fig.4 show the projections and the coordinates of special points and curves of them are attached.

*Remark 4:* For a given accuracy, precision, or recall, NI is not fixed (except some odd points) and its span is bonded by the special cases of NI.

*Remark 5:* For a given NI, accuracy, precision, and recall are not fixed (except some odd points) and their spans are not always in continuous domains, which are bonded by the special cases of NI.

Since accuracy is the combining of precision and recall (as in (11)), for the normal case (Case 9), NI can be written as the nonlinear function of precision with recall (as in (12)).

$$A = \frac{2PRw_1 + Pw_2 - Rw_1}{P(w_1+w_2)} \quad (11)$$

$$NI = \frac{1}{H(T)} \left\{ \log_2(w_1+w_2) + \frac{1}{P \cdot (w_1+w_2)} \begin{bmatrix} w_1\log_2 P^P \\ +Pw_1\log_2(1-R)^{(1-R)} \\ +Rw_1\log_2(1-P)^{(1-P)} \\ -PR\log_2 w_1^{w_1} \\ -P\log_2 w_2^{w_2} \\ +\log_2(PRw_1+Pw_2-Rw_1)^{(PRw_1+Pw_2-Rw_1)} \\ -\log_2(Pw_1+Pw_2-Rw_1)^{(Pw_1+Pw_2-Rw_1)} \end{bmatrix} \right\}$$
(12)

Analogously, precision is the combining of false alarm and hitting rate (as in (13)). So for the normal case (Case 9), NI can also be written as the nonlinear function of false alarm with hitting rate (recall) (as in (14)).

$$P = \frac{Rw_1}{Rw_1 + Fw_2} \quad (13)$$

$$NI = \frac{1}{H(T)}\left\{ \log_2(w_1+w_2) + \frac{1}{(w_1+w_2)}\begin{bmatrix} w_1 \log_2 R^R \\ + w_1 \log_2 (1-R)^{(1-R)} \\ + w_2 \log_2 F^F \\ + w_2 \log_2 (1-F)^{(1-F)} \\ - \log_2 (Rw_1+Fw_2)^{(Rw_1+Fw_2)} \\ - \log_2 (w_1(1-R)+w_2(1-F))^{(w_1(1-R)+w_2(1-F))} \end{bmatrix} \right\} \quad (14)$$

Fig. 5 shows the 3D ideal map of (12), regardless of the feasible region of precision with recall. Fig. 6 shows the feasible region of precision with recall and it is not $R^2$, i.e. $Pre. \times Rec. \neq R^2$. So the actual 3D map of (12) is shown in fig. 7. Equations (15) - (17) are the coordinations of the special point and curves of Fig. 6.

$$\alpha_{AP} : \left(\frac{w_1}{w_1+w_2}, 1\right) \quad (15)$$

$$\Gamma_{\alpha_{RP1}} : R = \frac{Pw_2}{(1-P)w_1} \quad (16)$$

$$\Gamma_{\alpha_{RP2}} : R = \frac{P}{(1-P)w_1} \quad (17)$$

Fig. 8 shows the 3D ideal map of (14), and the feasible region of false alarm with hitting rate (recall) is $R^2$, i.e. $F.A. \times Rec. = R^2$. So Fig. 8 is also the 3D actual map of (14).

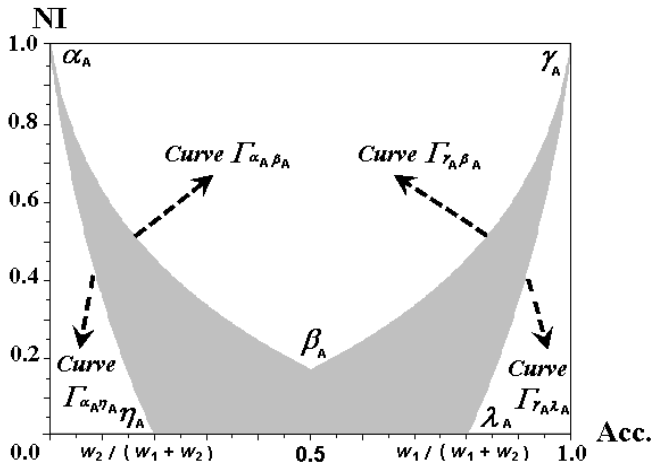

Fig. 2. NI/Acc. Nonlinear Relations Map when w1 ≥w2

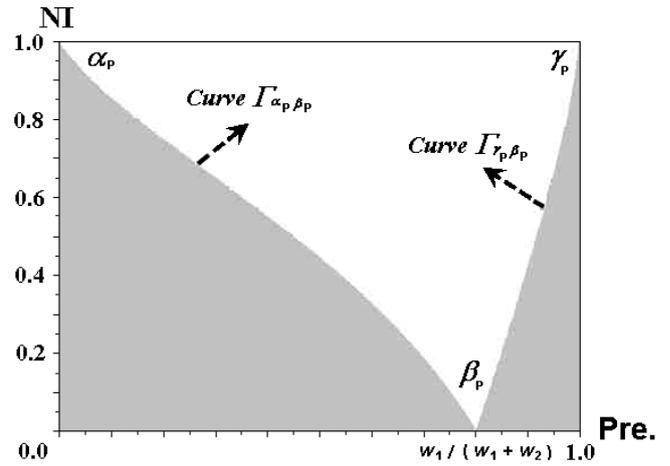

Fig. 3. NI/Pre. Nonlinear Relations Map when w1 ≥w2

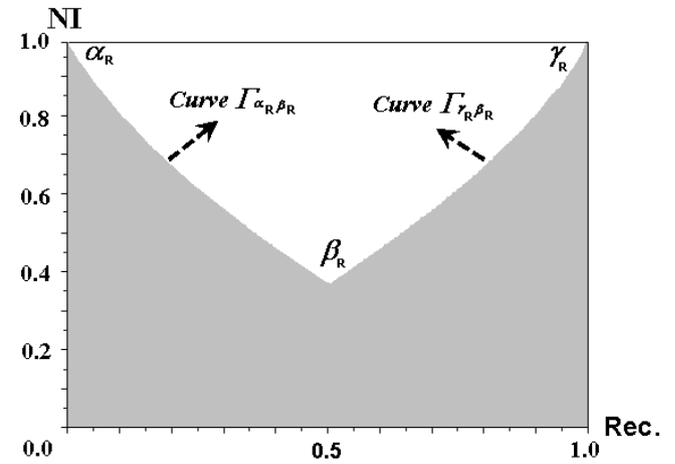

Fig. 4. NI/Rec. Nonlinear Relations Map when w1 ≥w2

$$\alpha_A : (0,1)$$

$$\beta_A : \left(0.5, \frac{-\frac{1}{2}+\frac{3}{2}\log_2(w_1+w_2)-\frac{w_2}{w_1+w_2}\log_2(w_2)-\frac{\frac{3}{2}w_1+\frac{1}{2}w_2}{w_1+w_2}\log_2\left(\frac{3}{2}w_1+\frac{1}{2}w_2\right)}{H(T)}\right)$$

$$\Leftrightarrow \Gamma_{\alpha_A \beta_A} = \Gamma_{\gamma_A \beta_A}$$

$$\gamma_A : (1,1)$$

$$\eta_A : \left(\frac{w_2}{w_1+w_2}, 0\right)$$

$\lambda_A : \left( \dfrac{w_1}{w_1+w_2}, 0 \right)$

$\Gamma_{\alpha_A \beta_A}$ : NI Case 5, $0 < A \le 0.5$

$\Gamma_{\gamma_A \beta_A}$ : NI Case 8, $0.5 \le A < 1$

$\Gamma_{\alpha_A \eta_A}$ : NI Case 6, $0 < A \le \dfrac{w_2}{w_1+w_2}$

$\Gamma_{\gamma_A \lambda_A}$ : NI Case 7, $\dfrac{w_1}{w_1+w_2} \le A < 1$

$\alpha_P : (0,1)$

$\beta_P : \left( \dfrac{w_1}{w_1+w_2}, 0 \right) \Leftrightarrow \Gamma_{\alpha_P \beta_P} = \Gamma_{\gamma_P \beta_P}$

$\gamma_P : (1,1)$

$\Gamma_{\alpha_P \beta_P}$ : NI Case 6, $0 < P \le \dfrac{w_1}{w_1+w_2}$

$\Gamma_{\gamma_P \beta_P}$ : NI Case 8, $\dfrac{w_1}{w_1+w_2} \le P < 1$

$\alpha_R : (0,1)$

$\beta_R :$

$\left( 0.5, \dfrac{-\dfrac{w_1}{2(w_1+w_2)} + \log_2(w_1+w_2) - \dfrac{w_1}{2(w_1+w_2)} \log_2(w_1)}{H(T)}, \dfrac{-\dfrac{\tfrac{1}{2}w_1+w_2}{w_1+w_2} \log_2\left(\tfrac{1}{2}w_1+w_2\right)}{H(T)} \right)$

$\Leftrightarrow \Gamma_{\alpha_R \beta_R} = \Gamma_{\gamma_R \beta_R}$

$\gamma_R : (1,1)$

$\Gamma_{\alpha_R \beta_R}$ : NI Case 6, $0 < R \le 0.5$

$\Gamma_{\gamma_R \beta_R}$ : NI Case 7, $0.5 \le R < 1$

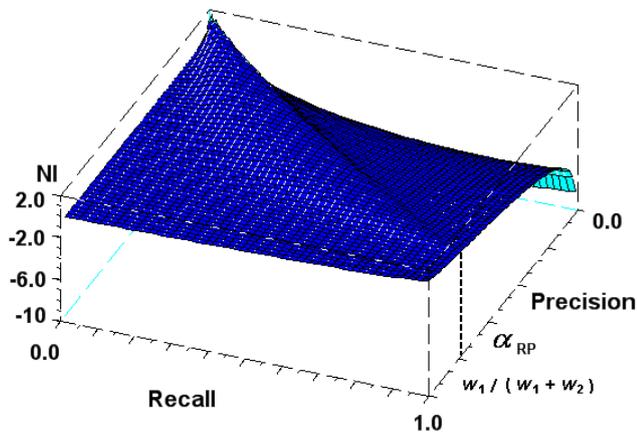

Fig. 5.  Ideal map of (12)

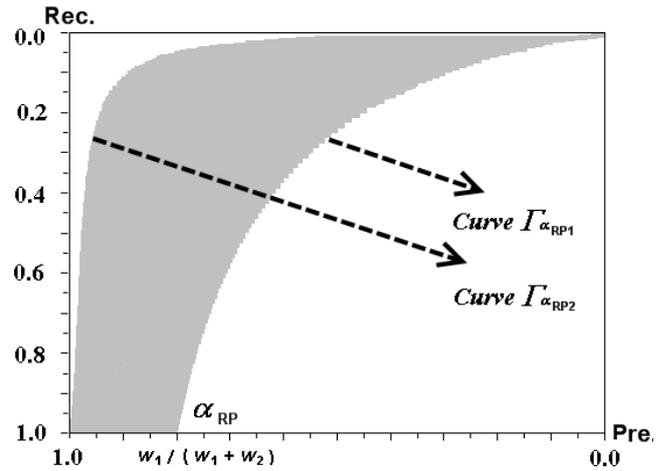

Fig. 6.  Feasible region of precision with recall

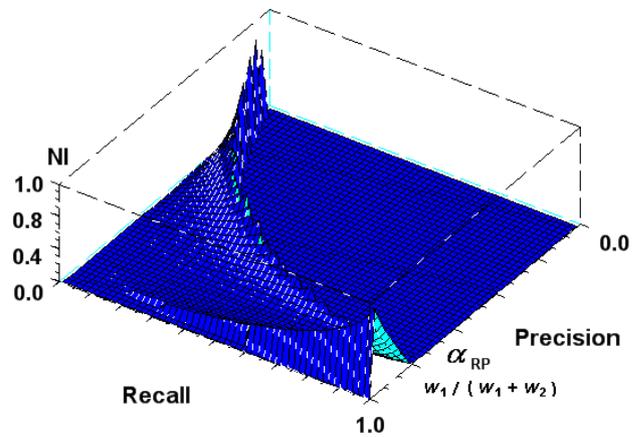

Fig. 7.  Actual map of (12)

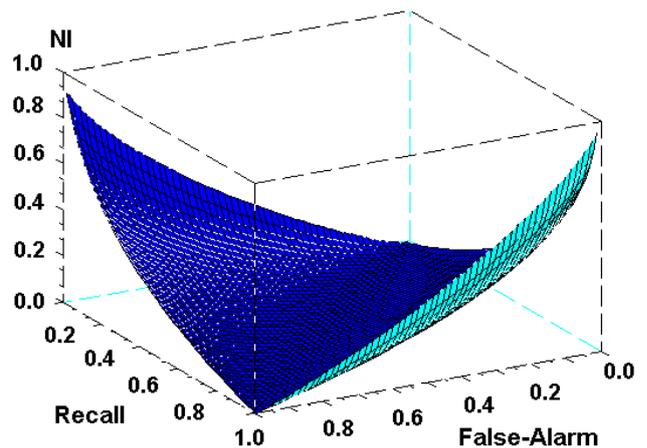

Fig. 8.  Ideal and actual map of (14)

## IV. EVALUATION

Being a novel, comprehensive and well-normalize criterion, many properties of NI are still unclear. But one thing is certain that NI is a very helpful complement for other criteria. When we compare different models on the basis of any criterion alone, we might meet the cases that the evaluation values are equal. So it is safer to evaluate different models with more than one criterion.

TABLE II
EXAMPLES IN BINARY CLASSIFICATONS (BASED ON EXAMPLE 1 DATA)

| Model | TP | FP | TN | FN | Accuracy | Precision | Recall | NI |
|---|---|---|---|---|---|---|---|---|
| $M_1$ | 25 | 5  | 45 | 25 | 0.7  | 0.8333 | 0.5  | 0.1468 |
| $M_2$ | 30 | 10 | 40 | 20 | 0.7  | 0.75   | 0.6  | 0.1245 |
| $M_3$ | 15 | 5  | 45 | 35 | 0.6  | 0.75   | 0.3  | 0.0468 |
| $M_4$ | 15 | 45 | 5  | 35 | 0.2  | 0.25   | 0.3  | 0.2958 |
| $M_5$ | 12 | 26 | 24 | 38 | 0.36 | 0.3158 | 0.24 | 0.0611 |
| $M_6$ | 26 | 12 | 38 | 24 | 0.64 | 0.6842 | 0.52 | 0.0611 |

*Definition 4: Model evaluation with NI and Accuracy.*

For binary classifiers $M_1$ and $M_2$, define the following model selection scheme:
1) If the NI of M1 is bigger than that of M2 and the accuracy of M1 is bigger than 0.5, M1 is chosen;
2) If the NI of M1 is bigger than that of M2 and the accuracy of M1 is smaller than 0.5, -M1 is chosen.;
3) If the NI of M1 is equal to that of M2 and the accuracy of M1 is bigger than that of M2, M1 is chosen.

*Example 1:* The total samples are 100. Half of them belong to the Positive class and the rest belong to the Negative class. $M_1 \sim M_6$ are six different classifiers for them and their accuracy, precision, recall, and NI are listed in Table II.

*Remarks on Example 1:*
1) Any criterion will meet the case that its values for different model are equal (shaded areas in Table II) and it can not make a model selection on its own. So it is safer to evaluate different models with more than one criterion;
2) Classifiers $M_4$ and $M_5$ satisfy the item 2 of Definition 4, so consider the models - $M_4$ and - $M_5$ for model evaluation;
3) The accuracy, precision, recall, and NI of - $M_4$ are 0.8, 0.75, 0.7, 0.2958;
4) The accuracy, precision, recall, and NI of - $M_5$ are 0.64, 0.6842, 0.76, 0.0611;
5) The evaluation result of the six different models is:
$$- M_4 > M_1 > M_2 > - M_5 > M_6 > M_3$$

## V. DISCUSSION

How to choose the model evaluation criteria is still an open problem in machine learning. As a preliminary study, the main objective of this paper is not to prove what criterion is the best for classifier evaluation but to propose a new information-based criterion, NI, and to reveal that NI integrates conventional performance-based criteria, accuracy, precision, and recall, in a nature way (it is actually a concordant power-exponent function of them), and it can also be expressed with respect to precision and recall, or to false alarm and hitting rate (recall).

Performance-based criteria and information-based criteria are two different categories for model evaluation and their relations are still on the research. In this paper, we demonstrate how to apply an information-based criterion (NI) with accuracy to model evaluation, but many properties of NI are still unclear and further investigations are needed to be done for analyzing the relations between other information-based criteria and NI.

ACKNOWLEDGMENT

The authors wish to express their thanks to Yu-Jiu Yang and Shuang-Hong Yang for their enlightening discussions and constructive suggestions.